\documentclass[runningheads]{llncs}
\usepackage[T1]{fontenc}
\usepackage{graphicx}
\usepackage[ruled,linesnumbered]{algorithm2e}
\usepackage{relsize}
\usepackage{subfigure}
\usepackage{amsmath}
\usepackage{amssymb}
\usepackage{booktabs}
\usepackage{makecell}

\begin{document}
\title{Anomaly Correction of Business Processes Using Transformer Autoencoder}
\author{Ziyou Gong\inst{1,2}\orcidID{0000-0003-1724-1364} \and
Xianwen Fang\inst{1,2}\orcidID{0000-0001-8531-7215} \and
Ping Wu\inst{1}\orcidID{0009-0009-6631-1032}}
\authorrunning{Z. Gong et al.}
\institute{Anhui University of Science and Technology, Huainan, China \\
\email{gzy409338163@163.com} \and
Anhui Province Engineering Laboratory for Big Data Analysis and Early Warning Technology of Coal Mine Safety, Huainan, China}
\maketitle              % typeset the header of the contribution
\begin{abstract}
Event log records all events that occur during the execution of business processes, so detecting 
and correcting anomalies in event log can provide reliable guarantee for subsequent process analysis. 
The previous works mainly include next event prediction based methods and autoencoder-based methods. 
These methods cannot accurately and efficiently detect anomalies and correct anomalies at the same time, 
and they all rely on the set threshold to detect anomalies. To solve these problems, we propose a 
business process anomaly correction method based on Transformer autoencoder. By using self-attention 
mechanism and autoencoder structure, it can efficiently process event sequences of arbitrary length, 
and can directly output corrected business process instances, so that it can adapt to various scenarios. 
At the same time, the anomaly detection is transformed into a classification problem by means of 
self-supervised learning, so that there is no need to set a specific threshold in anomaly detection. 
The experimental results on several real-life event logs show that the proposed method is superior to 
the previous methods in terms of anomaly detection accuracy and anomaly correction results while ensuring 
high running efficiency.

\end{abstract}
\section{Introduction}
\label{sec 1}
Anomaly detection in business processes is an important means to guarantee the correct execution and timely correction of business processes. At the same time, the execution of business processes will record event logs in the information system, on which many process analysis methods depend, and the repair of low-quality event logs, i.e., anomaly correction, also facilitates the subsequent analysis of business processes \cite{van2022process,ko2023systematic}.

Previous anomaly detection methods of business process based on predictive process monitoring first predict the next event of business process through machine learning method, and set a threshold to determine whether the business process instance is anomalous by whether it exceeds or falls below this threshold \cite{nolle2022binet,lee2022analysis}. Most of the machine learning models for predicting the next event are LSTM \cite{evermann2017predicting,tax2017predictive,camargo2019learning}, but the feature extraction capability of LSTMs and the parallelism of model training are difficult to satisfy all business process instances. And this methods requires iteratively predicting the next event for each process instance, which is less efficient in offline scenarios. 

Autoencoder-based anomaly detection and anomaly correction trains a denoising autoencoder by injecting noise into the normal event log, and ensures that the output is close to the original normal event log by minimizing the event log reconstruction error \cite{nolle2016unsupervised,huo2021graph,krajsic2021variational,nguyen2019autoencoders}. The autocoder-based approach inputs a complete business process instance with anomalies and outputs a complete business process instance without anomalies, which can be more efficient compared to predicting the next event. However, previous autoencoder-based methods can only handle a limited number of anomalies and all of them set the threshold and thus discriminate the anomalies by averaging the reconstruction error, which relies heavily on the model to learn the difference between the characteristics of normal and anomalous cases.

In recent years Transformer \cite{vaswani2017attention} has shown excellent performance in the field of natural language processing and computer vision. Compared with LSTM, it can process all elements of the input sequence at the same time, so the model has higher parallelization and is more suitable for large-scale scenarios. Therefore, Transformer have also been noticed by researchers in the field of process mining \cite{philipp2020predictive,bukhsh2021processtransformer}. The mask pre-training models developed on this basis utilize Transformer's excellent feature extraction capability to achieve the best performance in various tasks in the fields of natural language processing and computer vision \cite{devlin2018bert,he2022masked}. It enhances the feature learning capability of the model by predicting the masked parts of input sentences by context.

In this paper, we propose a business process anomaly detection and anomaly correction method based on Transformer autoencoder (TransformerAE), which utilizes the Transformer autoencoder and self-supervised learning on original event logs. Compared with previous methods, we transform the anomaly detection problem into a classification problem so that it does not need to set a threshold for anomaly determination, and leverages transformer's structure to combine it with anomaly correction tasks. Evaluation results of anomaly detection and anomaly correction task on 4 real-life event logs comparing with baseline methods show that our method has a better balance of detection and correction results and running speed. 

The rest of the paper is organized as follows: Section \ref{sec 2} focuses on some relate work on business process anomaly detection and business process anomaly correction. Section \ref{sec 4} focuses on the Transformer autoencoder-based approach for business process anomaly detection and anomaly correction. Section \ref{sec 5} describes the data used for the experiments and the evaluation and comparison of the methods in this paper. Section \ref{sec 6} summarizes the paper as well as provides an outlook for future work.

\section{Related work}
\label{sec 2}

Business process anomaly detection methods are usually closely related to business process anomaly correction methods, and a lot of previous work has been proposed, which can be mainly divided into classical methods and machine learning methods. Classical methods mainly focus on statistical analysis of event logs and manual feature extraction. For business process anomaly detection, Böhmer et al. \cite{bohmer2016multi} proposed a multi-angle anomaly detection method, which can deal with unexpected situations of process model execution events and combine multiple events to detect collective anomalies. Sarno et al. \cite{sarno2020anomaly} proposed an anomaly detection method based on fuzzy multi-attribute decision making and fuzzy association rule learning. Van Zelst et al. \cite{van2020detection} proposed a general event flow filter, which detects and removes infrequent behaviors from the event stream by constructing a probabilistic automaton set and dynamically updating it to filter anomalous events. Ko et al. \cite{ko2021detecting} propose a statistical lever-based process anomaly scoring method that uses three different methods to set anomaly detection thresholds. It can be integrated into an online framework to handle anomalies in an online event stream \cite{ko2022keeping}. For business process anomaly correction, Suriadi et al. \cite{suriadi2017event} proposed a pattern-based approach to record common event log quality problems, developed a set of components to describe event log quality problems as patterns, and used the pattern library to identify and repair event log quality problems. Fani Sani et al. \cite{sani2019repairing} used a probabilistic approach to identify anomalous behavior based on the behavioral context of the process, i.e. the sequence of fragments of activity that occurred before and after the potentially anomalous behavior, and then replaced them with behaviors that were more likely to occur in the context in which the anomalous behavior occurred. Liu et al. \cite{liu2021repairing} proposed a missing activity repair method based on activity sequence relationships in event logs, which used activity relationship matrix to represent event logs and cluster them.

Machine learning methods make feature extraction easier and make end-to-end anomaly detection and correction easier. For business process anomaly detection, Pauwels et al. \cite{pauwels2019anomaly} proposed a model that uses dynamic bayesian networks to model the normal behavior in log files and point out the root cause of the anomaly. Tavares et al. \cite{tavares2019leveraging} proposed an online anomaly detection method for business processes, which extracted case descriptors from event streams and applied density-based clustering techniques to detect outliers. Autoencoder-based approaches learn the representation of the underlying model to detect anomaly traces and abnormal activity \cite{nolle2016unsupervised,krajsic2021variational}, or set an anomaly detection threshold by averaging reconstruction losses \cite{nguyen2019autoencoders,huo2021graph}. Lee et al. \cite{lee2022analysis} solve the problem of anomaly detection of online events by proposing a method that adopts next stage activity prediction, which uses a machine learning model to predict the probability of the next activity and treats the unpredictable events as anomalies. Binet \cite{nolle2022binet} is a neural network architecture for anomaly detection of multi-view business process event logs, and a set of heuristic algorithms for automatically setting the threshold of anomaly detection algorithms. For business process anomaly correction, \cite{nguyen2019autoencoders} proposed an autoencoder-based event log repair method to predict missing or incorrect events by training the autoencoder on the event log. Nolle et al. \cite{nolle2020deepalign} proposed a multi-view process anomaly correction method based on recurrent neural networks and bidirectional beam search, using two trained recurrent neural networks to predict event sequence from left to right and calculate event sequence alignment from right to left, as a way to detect and correct event sequence anomalies.

Few previous works combine anomaly detection and anomaly correction of business processes together, and cannot achieve end-to-end anomaly detection and anomaly correction. Moreover, most anomaly detection methods rely on specific threshold Settings, and different thresholds need to be designed for different event logs.

\section{Methodology}
\label{sec 4}
\begin{figure*}
	\centering
	\includegraphics[scale=.2]{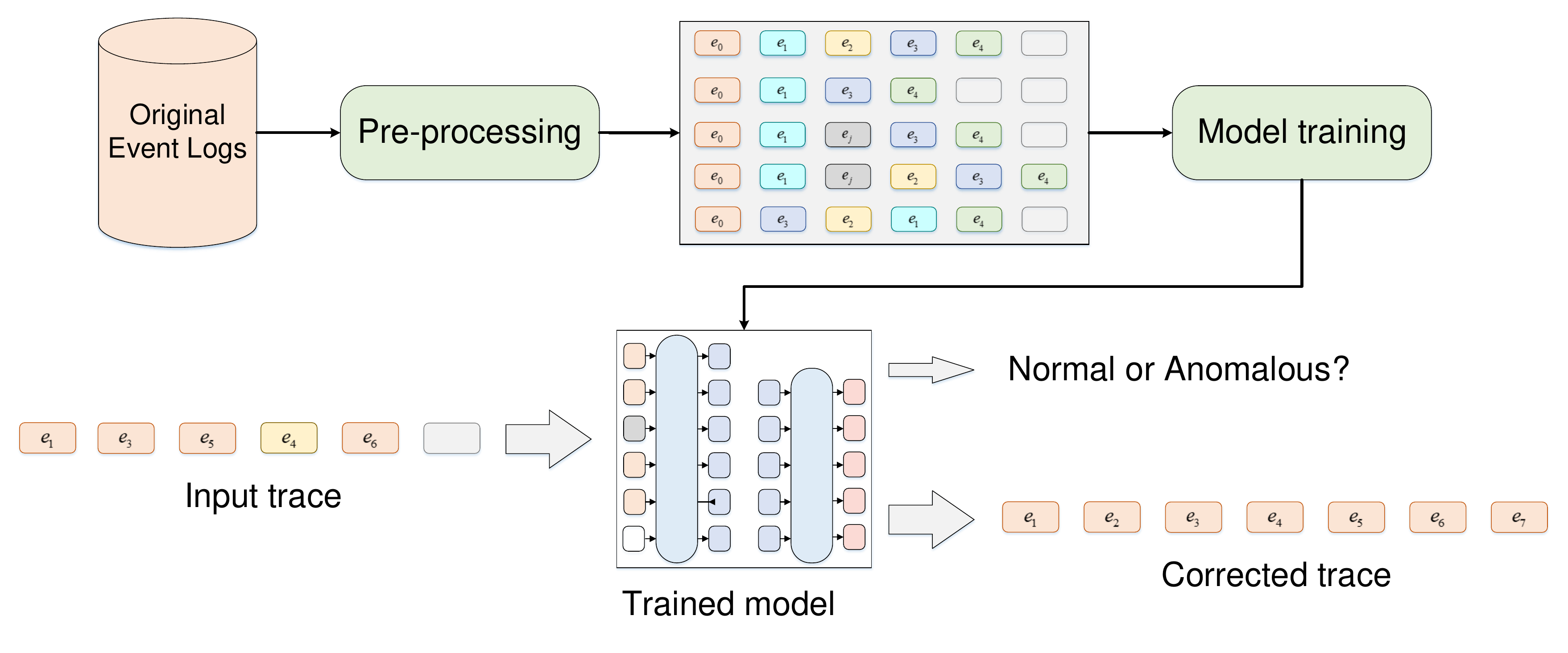}
	\caption{Overview of proposed method.}
	\label{FIG:3}
\end{figure*}

In this section, we first give the overview of proposed methods. As shown in Fig. \ref{FIG:3}, the input of the model is an event log containing anomalies constructed from the original nomal event log. The procedure for creating an event log containing anomalies will be described in detail in Section \ref{subsec 4.1}. The model will classify each trace to determine whether it is an anomaly trace, and output the corrected input trace. The detailed model structure and training process will be presented in Section \ref{subsec 4.2}.

\subsection{Pre-processing}
\label{subsec 4.1}
We first introduce some basic concepts of event logs. Let $A$ be the set of $h$ distinct activities, $T$ be the time domain, and $C$ be the set of case identifiers. An event is a tuple $\varepsilon=\left(c,a,t\right)$ where $c \in C$, $a \in A$, $t \in T$. A trace $\sigma=\langle {{\varepsilon}_{1}}, {{\varepsilon}_{2}}, \ldots, {{\varepsilon }_{m}}\rangle$ is the execution trajectory of an instance of a business process. It is a finite sequence of $m$ distinct events with the same case identifier that is non-decreasing in time. So the event log $L=\{{{\sigma}_{1}}, {{\sigma}_{2}}, ..., {{\sigma}_{n}}\}$ is a set of $n$ different traces of all records of business process execution.

Anomaly detection in business processes requires event logs with labeled traces (normal or anomaly), which is usually not available in practice. Therefore, previous research usually inject different types of anomalies in the event log and label each trace \cite{ko2021detecting,lee2022analysis,bohmer2016multi}. We redefines 6 types of anomalies based on the previous research \cite{ko2021detecting,lee2022analysis,bohmer2016multi} which are Missing, Skip, Replace, Insert, Early and Late. The 6 types of anomalies are described as follows:

\begin{itemize} 
	\item \textbf{Missing}: An event in the process instance is recorded in the event log, but the activity identifier is missing. 
	\item \textbf{Skip}: An event in the process instance is skipped. 
	\item \textbf{Replace}: An event in the process instance is replaced with an arbitrary other event. 
	\item \textbf{Insert}: Insert a random event into the process instance. 
	\item \textbf{Early}: an event in the process instance occurs early.
	\item \textbf{Late}: an event in the process instance is postponed.
\end{itemize}

Then we adopt the masking operation similar to that in BERT to inject at least one kind of anomaly into a certain proportion of traces, which is determined by $r_{case}$, and the maximum number of anomaly categories is determined by $r_{act}$. Since Skip and Insert anomalies will change the length of traces, the position of each injected anomaly will be restricted in order to prevent some injected anomalies from affecting other anomalies. A limit on the maximum length of the input sequence is required to ensure a same length of the input sequence, which is also determined by the proportion $r_{act}$ and the anomaly type, i.e., the maximum length of the input sequence is all anomalies injected in the maximum length traces are of Insert anomaly type. In addition, it is known from the log behavior profile \cite{weidlich2010process} that the traces created by some of the above anomalies may be normal. For example, if two events have a interleaving order relation, the injection of the Early and Late anomalies results in an exchange of the order of these two events in the business process instance that does not result in an anomaly. Therefore for this case it needs to be labeled as normal. The anomaly trace detection problem will be transformed into a classification problem, that is, let the classification model learn the difference between normal traces and anomalous traces.

\subsection{Anomaly Detection and Anomaly Correction with TransformerAE}
\label{subsec 4.2}
\begin{figure*}[htbp]
	\centering
	\makebox[\textwidth][c]{\includegraphics[scale=.32]{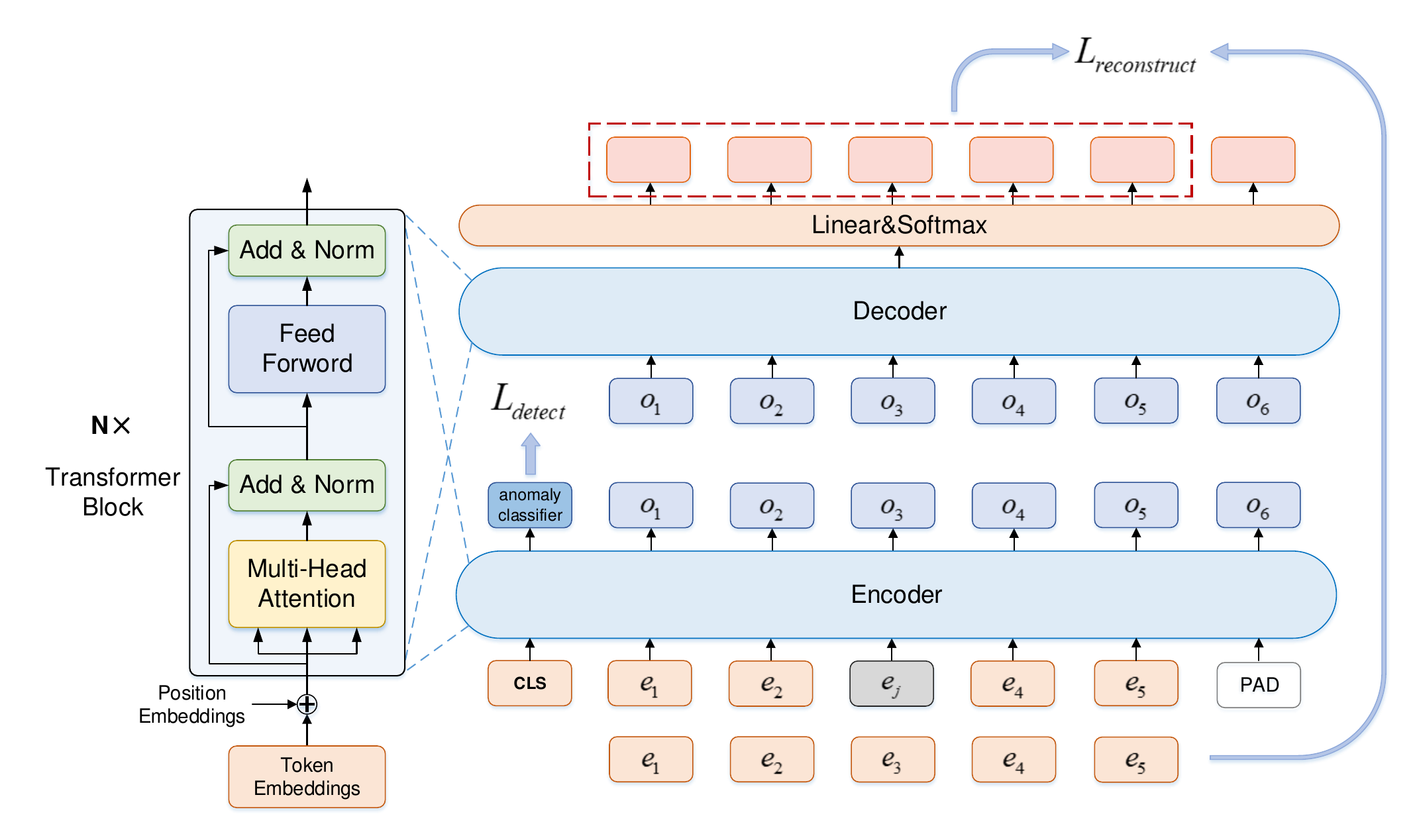}}
	\caption{Architecture of TransformerAE.}
	\label{FIG:6}
\end{figure*}

The structure of the TransformerAE model proposed in this paper is shown in Fig. \ref{FIG:6}. The model consists of two parts, namely Encoder and Decoder, both Encoder and Decoder are stacked based on Transformer Block, and the Transformer Block used is the Encoder of the original Transformer. Suppose that the input sequence is $X$, the multi-head attention mechanism on $X$ is computed as follows:

\begin{equation}
	\begin{aligned}
		\mathrm{MultiHeadAttention}(X)&=\mathrm{Contact}(\mathrm{head}_1,\dots,\mathrm{head}_n)W^O\\
		\text{where} \ \mathrm{head}_i&=\mathrm{SelfAttention}(X)\\
		&=\mathrm{Attention}(X,X,X)\\
		&=\mathrm{softmax}(\frac{XW_i^Q(XW_i^K)^T}{\sqrt{d_k}})XW_i^V
	\end{aligned}
\end{equation}

Where $W_i^Q$, $W_i^K$, $W_i^V$ and $W_i^O$ is the learnable weights. Then the Transformer Block is defined as follows:

\begin{equation}
	\begin{aligned}
		Z=\mathrm{Norm}(\mathrm{MultiheadAttention}(X)+X)\\
		\mathrm{Transformer \ Block}=\mathrm{Norm}(\mathrm{FFN}(Z)+Z)
	\end{aligned}
\end{equation}

Where Norm is the normalization layer and FFN is the feed forward neural network. Based on this, TransformerAE is composed of transformer block encoder and decoder, and input sequence first will be processed by an embedding layer consisting of a learnable token embedding and a learnable position embedding before being fed into encoder and decoder:

\begin{equation}
	\begin{aligned}
		Y=\mathrm{Embedding}(X&)+\mathrm{Position Embedding}(X)\\
		\mathrm{Encoder}(Y)=\mathrm{Transformer \ Block}(\dots & \mathrm{Transformer \ Block}(
		\mathrm{Transformer \ Block}(Y)))\\
		\mathrm{Decoder}(Y)=\mathrm{Transformer \ Block}(\dots & \mathrm{Transformer \ Block}(
		\mathrm{Transformer \ Block}(Y)))\\
		\mathrm{TransformerAE}&=\mathrm{Decoder}(\mathrm{Encoder(X)})
	\end{aligned}
\end{equation}

The trace is represented as a sequence of events after preprocessing, then the classification token "[CLS]" is added to the head of the event sequence and the token "[PAD]" is added to the tail of the event sequence to fill the event sequence into the same length. The event sequence is encoded by one-hot encoding and then fed into Encoder, which obtains the embedding information of the event sequence through token embedding and position embedding. Decoder's input also consists of token embedding and position embedding, the difference from Encoder is that the first position of the sequence does not have the classification token "[CLS]". Decoder's output consists of the linear layer and Softmax activation function to obtain the classification result of each element of the event sequence, i.e. corrected trace. In this way, the corrected trace can be output at one time, which is more efficient than the autoregressive output.

Since there is no way to know whether there is any anomaly in the event log in the existing dataset, it is necessary to inject random anomalies into the original event log which is assumed to be free of anomalies in order to simulate possible anomalies in the real world, so as to test the performance of the trained model. Assuming an original event log $L=\{{{\sigma }_{1}},{{\sigma }_{2}},...,{{\sigma }_{n}}\}$ with no anomalies, and injecting anomalies into it to get an event log ${L}'=\{{{{\sigma }'}_{1}},{{{\sigma }'}_{2}},...,{{{\sigma }'}_{n}}\}$ containing anomalous events, then the anomaly correction task in this paper is to make $\hat{L}=\mathrm{Decoder}(\mathrm{Encoder}({L}'))$ as close as possible to $L$ by using an Encoder and a Decoder. The anomaly detection task is also based on the output of the Encoder, i.e., defining a detection function $f$, there:

\begin{equation}
	\begin{aligned}
		f(\mathrm{Encoder}({{\sigma }_{i}})[0])=\mathrm{sigmoid}(\mathrm{Linear}(\mathrm{Encoder}({{\sigma }_{i}})[0]))
	\end{aligned}
\end{equation}

\begin{flushleft}
Where $[0]$ denotes the first element of the taken sequence, corresponding to the first element of the input sequence, i.e., the classification token "[CLS]", $\mathrm{Linear}$ denotes the linear mapping, and $\mathrm{sigmoid}$ is the activation function.
\end{flushleft}

During model training, the training samples created from the original traces are input into Encoder, then the output of the position corresponding to the classification token "[CLS]" is fed into an anomaly classification layer to get the classification result of the input trace, and the loss function for the binary classification of normal and anomalies is ${{L}_{detect}}=-y\log \hat{y}-(1-y)\log (1-\hat{y})$. The rest of the output of Encoder is then fed into Decoder. For the correction of anomaly traces, unlike the autocoder-based anomaly correction which uses a squared loss function, here the cross-entropy loss is computed from the Decoder's output and the original input sequence, i.e., the reconstructed loss function is ${{L}_{reconstruct}}=-\sum\nolimits_{i=1}^{h+2}{{{y}_{i}}\log {{{\hat{y}}}_{i}}}$, where $h$ is the total number of activity categories ,because the "[CLS]" and "[PAD]" tokens are added, and the total loss is $L={{L}_{detect}}+{{L}_{reconstruct}}$. Since correcting some specific anomalies changes the length of the event sequence, the complete input sequence and output sequence need to be considered in computing the reconstruction loss, which introduces unnecessary filling of tokens, but this effect was found to be negligible in the experiments.

\section{Evaluation}
\label{sec 5}
\subsection{Experimental data and experimental setup}
\label{subsec 5.1}
For the experiments, we use four real-life event log datasets from different domains which are available at 4TU.ResearchData (https://data.4tu.nl/). All the event log statistics are shown in Table \ref{tab 2}.

\begin{table}[h]
	\caption{Event log dataset statistics.}
	\label{tab 2}
	\centering
	\begin{tabular}{p{2cm}<{\centering}p{2cm}<{\centering}p{2cm}<{\centering}p{2cm}<{\centering}p{2cm}<{\centering}}
		\toprule
		Event logs & \makecell{Num. of \\ trace} & \makecell{Num. of \\ activity} & \makecell{Avg. \\ case length} & \makecell{Max. \\ case length} \\
		\midrule
		BPIC2012W & 9658 & 6 & 7 & 79      \\
		BPIC2012 & 13087 & 24 & 20 & 175      \\
		BPIC2013in & 7554 & 7 & 8 & 123      \\
		BPIC2017 & 31509 & 26 & 38 & 180      \\
		\bottomrule
	\end{tabular}
\end{table}

In the experiment, 80\% of the entire event log is used for training, and the remaining is used to test the effect of anomaly classification and anomaly correction. During training, case anomaly percentage is set to 0.5 for class balance, and event anomaly percentage is set ot 0.3, 0.5 and 0.7 respectively (the three anomaly pecentage only represent fewer, balanced, and more anomalous events, and the best results can be obtained through more fine-tuning in real-life scenario), and for testing, the case anomaly pecentage is selected as 0.1, 0.3, 0.5, and 0.7, while the event anomaly percentage is at 0.1, 0.3, 0.5 and 0.7 on the basis of increasing the number of anomalous events for 1 and 2 cases, used to detect the performance of all methods in the most extreme cases, note that some event logs or part of the maximum length of the trace is small, 1 or 2 events may also account for the proportion of the total length of the trace is more than 0.1, so increase the number of anomalous events for the case of 1 and 2 in order to try to simulate all the cases in reality.

We compare our method with four neural network architectures, including BINet \cite{nolle2022binet}, Autoencoder (AE), LSTM Autoencoder (LSTMAE), and BERT. BINet is a time-step prediction-based anomaly detection method, which has three versions respectively. Since no other event attribute except activity is used in this paper, only the BINetv1 is compared. The anomaly detection method based on AE and LSTMAE is based on the  \cite{nguyen2019autoencoders}, which sets a threshold of anomaly detection by the average reconstruction loss of the whole event log, and then discriminates whether the trace is anomalous or not, and the reconstruction results of AE and LSTMAE are corrected traces. In this paper, the hyperparameters of the LSTMAE are adjusted (the dimension of the hidden layer is adjusted to 100, and the number of the LSTM layers is adjusted to 2), so as to achieve a better performance. The BERT-based method replaces the decoder of TransformerAE with a fully connected neural network, and the procedure of anomaly detection is the same as TransformerAE. The hyperparameter settings for BERT and TransformerAE in Transformer Block are shown in Table \ref{tab 3}. 

\begin{table*}[h]
	\caption{Transformer Block hyperparameter settings.}
	\label{tab 3}
	\centering
	\begin{tabular}{p{7.5cm}<{\centering}p{1cm}<{\centering}p{3cm}<{\centering}}
		\toprule
		& BERT & TransformerAE \\
		\midrule
		Num. of encoder attention heads & 8 & 8 \\
		Num. of decoder attention heads & - & 8 \\
		Num. of Transformer Block layers of the encoder & 2 & 2 \\
		Num. of Transformer Block layers of the decoder & - & 2 \\
		Dimension of FFN & 64 & 64 \\
		\bottomrule
	\end{tabular}
\end{table*}

\subsection{Evaluation metrics}
\label{subsec 5.3}
For the results of case-level anomaly detection, we use the F-score for case classification to evaluate the proposed method. At the event level, the same evaluation metric as for anomaly correction are used because some anomalies lead to changes in the length of the traces and the position of the events, which makes it difficult to accurately measure them by the accuracy of event prediction. The position and type of anomalous events are determined by comparing the corrected traces with the input traces.

For anomaly correction results, some works usually only repairs missing and replaced activities so only the prediction accuracy of missing and replaced activities is evaluated. For real-life anomaly event logs, however, not only missing and replaced activities are included, but also missing events or changes in event locations and redundant event records may be included, so each complete trace in the event log needs to be compared during anomaly correction. In view of this, we consider using similarity based on Damerau-Levenstein distance (edit distance) to evaluate the results of anomaly correction, i.e., we compute the similarity between the anomalous traces and the corrected traces.

\begin{equation}
	\begin{aligned}
		{{S}_{DL}}({{s}_{1}},{{s}_{2}})=1-\frac{\mathrm{DL \_ distance}({{s}_{1}},{{s}_{2}})}{\max (\mathrm{len}({{s}_{1}}),\mathrm{len}({{s}_{2}}))}
	\end{aligned}
\end{equation}

\begin{flushleft}
where $\mathrm{DL \_ distance}({{s}_{1}},{{s}_{2}})$ is the edit distance, and $\mathrm{len}(s)$ is the length of the sequence $s$. A similarity of 0 means completely different and a similarity of 1 means completely identical.
\end{flushleft}

\subsection{Experimental results}
\begin{figure*}[htbp]
	\centering
	\makebox[\textwidth][c]{\includegraphics[scale=.34]{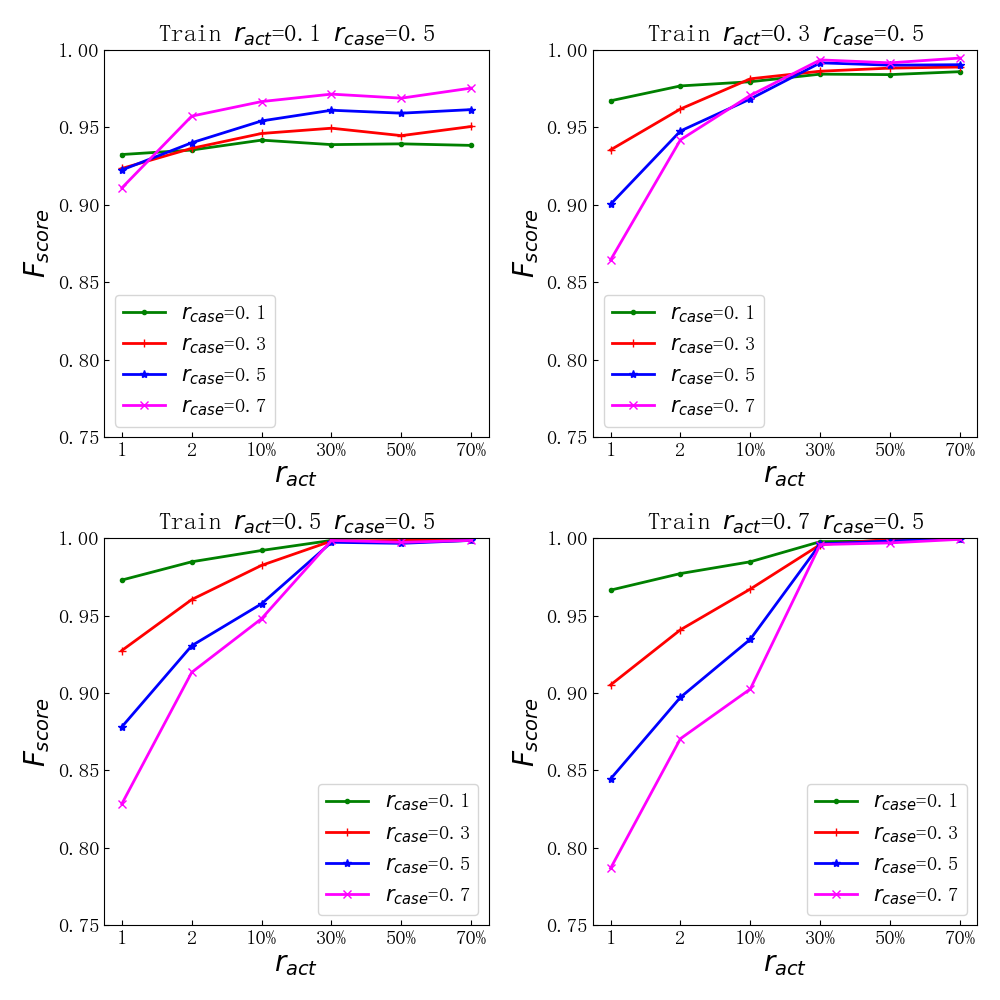}}
	\caption{Results of anomalous case classification task on BPIC2012.}
	\label{FIG:8}
\end{figure*}

The experimental results of the anomalous case classification on the event log BPIC2012 are shown in Fig. \ref{FIG:8}, each subimage corresponds to the event anomaly proportion ${{r}_{act}}$ is 0.1, 0.3, 0.5 and 0.7 respectively during traing. Where each curve represents the different case anomaly proportion ${{r}_{case}}$ at the time of testing and the horizontal coordinate is the different event anomaly proportion ${{r}_{act}}$ at the time of testing. From the figure, it can be seen that injecting different event anomaly proportion during training has a greater impact on the results, and for the lower the number of anomalous events, the worse the classification effect is, so a lower event anomaly proportion is needed to select during training.

\begin{figure*}[htbp]
	\centering
	\makebox[\textwidth][c]{\includegraphics[scale=.34]{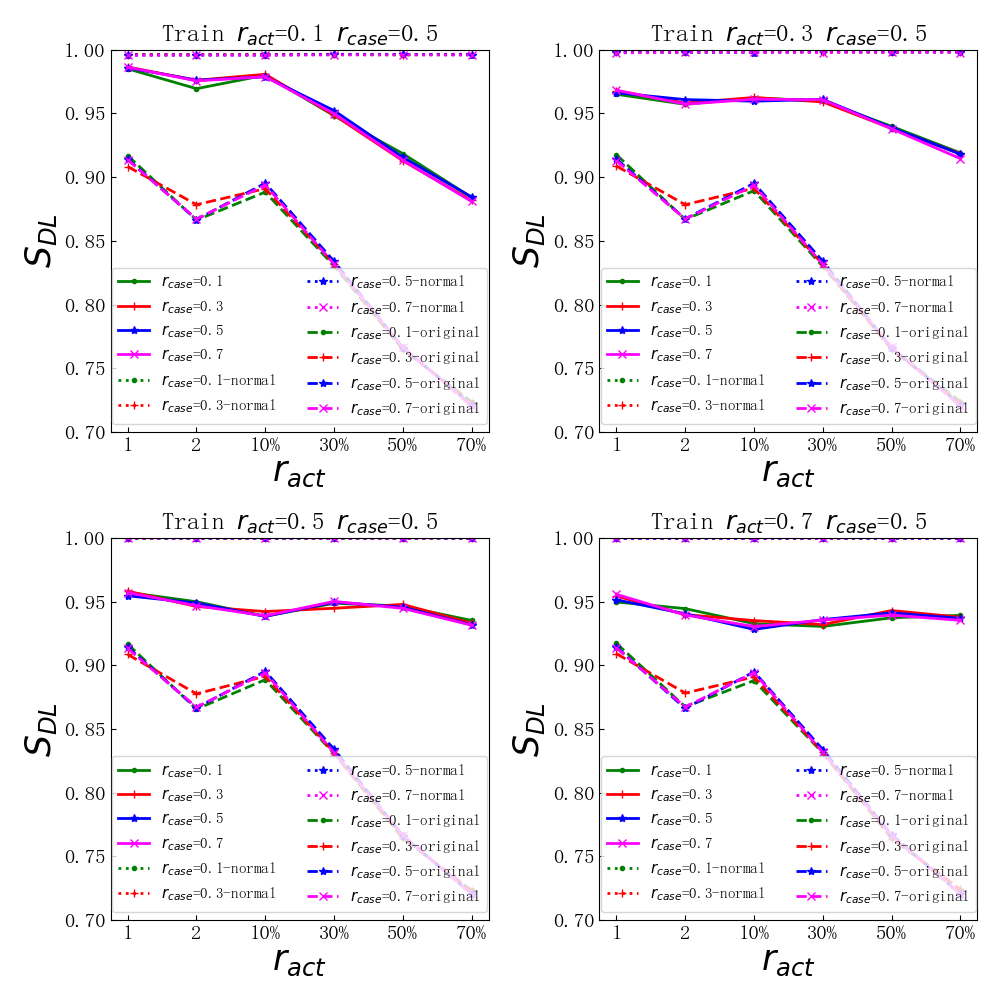}}
	\caption{Results of anomalous correction task on BPIC2012.}
	\label{FIG:9}
\end{figure*}

The anomalous proportion setting in the experiment of anomaly correction is the same as the experiment of anomalous case classification, and the results are as follows Fig. \ref{FIG:9}. The normal (dotted dashed line) indicates the change to the traces originally labeled as normal, and it can be found that TransformerAE almost does not change the normal traces, i.e., it can distinguish the normal traces from the anomalous traces well, and at the same time the effect on the normal traces can be almost negligible even though the model may make a mistake in discriminating the normal traces. The original (short dashed line) indicates the similarity between the anomalous traces and the traces before anomalies are injected, and the solid line is the similarity between the corrected traces and the traces before anomalies are injected. This can be shown that TransformerAE has some ability to correct the anomalous traces, and the larger the event anomaly proportion is, the more obvious is the effect of correction. When the event anomaly proportion is 0.3 during training, the result of anomaly correction decreases more dramatically for higher event anomaly proportion (70\%) during testing, while the result of testing decreases less with event anomaly proportion of 0.5 and 0.7 for training. This shows that for the anomaly detection task and the anomaly correction task a balance needs to be found so that both achieve better results.

\begin{figure*}[htbp]
	\centering
	\makebox[\textwidth][c]{\includegraphics[scale=.31]{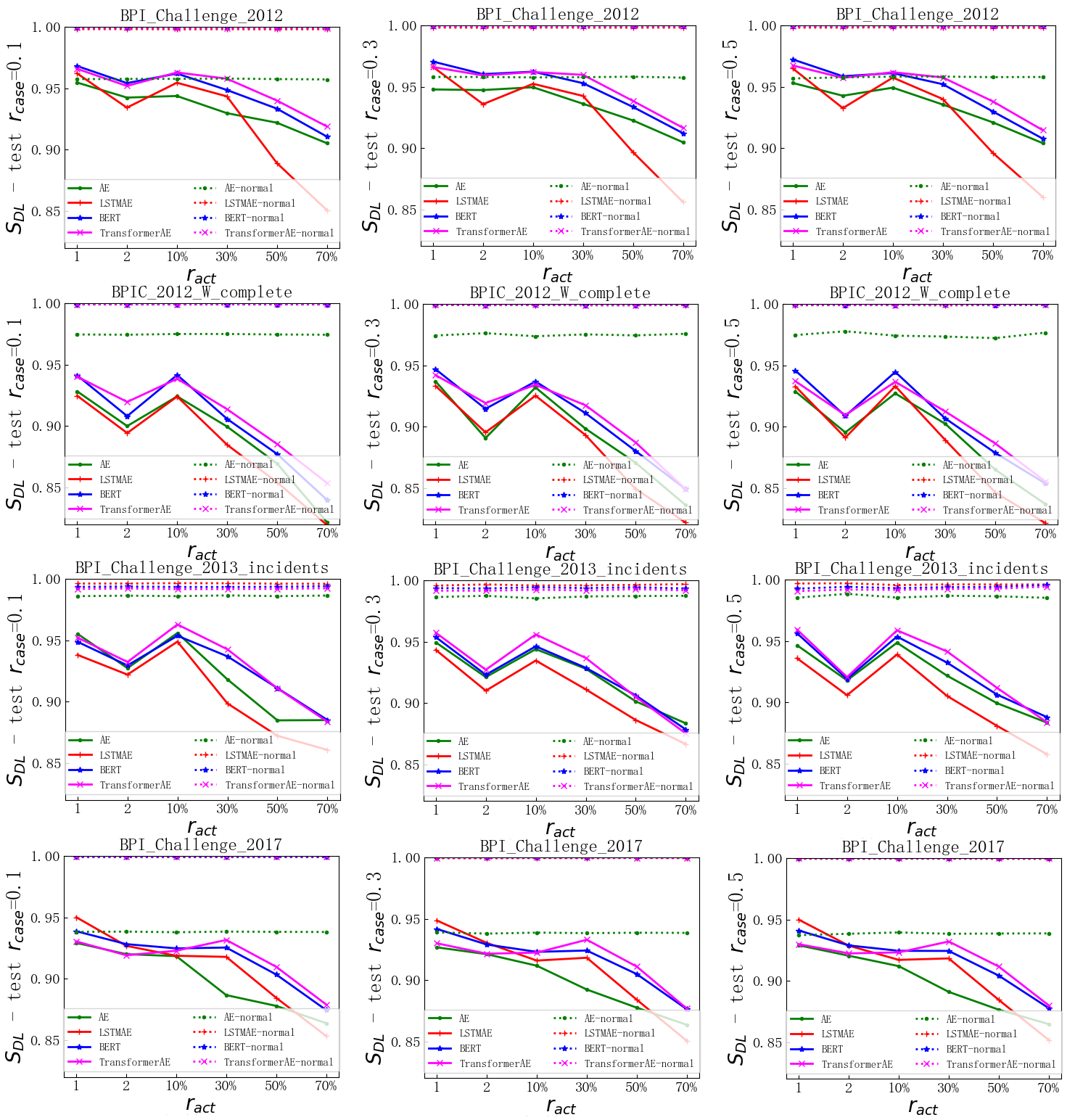}}
	\caption{Comparison of anomaly correction tasks with baseline methods.}
	\label{FIG:11}
\end{figure*}

The comparison with the baseline methods in the anomaly detection task is shown in the Table \ref{tab 4}, in order to make the experimental setting close to BINet, therefore the event anomaly proportion are set to 0.3 for both training and testing. It can be seen that BINetv1 shows similar performance on all logs and are lower than the autocoder-like methods, which may be due to the fact that BINet is designed to be used on the conditions that the anomalies are unknown in the original logs, and therefore the detection is less effective. In the anomaly detection task, the results of TransformerAE and BERT are basically the same, this is due to the fact that the anomaly detection part of TransformerAE is same to BERT. The results of AE are much worse than that of other models. Combined with the result in Fig. \ref{FIG:11}, this may be because AE changes both anomalous traces and normal traces, resulting in close reconstruction errors between normal traces and anomalous traces, so that it is hard to distinguish them well. The results of LSTMAE are better than AE because the feature learning ability of LSTM for event sequences is much better than that of fully connected neural networks. However, the threshold setting based on the average reconstruction error can only give slightly better results when the proportion of normal traces and anomalous traces is close to each other, and the normal traces and anomalous traces can't be well distinguished in many cases.

\begin{table*}[h]
	\caption{Comparison of anomaly detection tasks with baseline methods (F-score).}
	\label{tab 4}
	\centering
	\begin{tabular}{p{2cm}<{\centering}p{1.2cm}<{\centering}p{1cm}<{\centering}p{1.5cm}<{\centering}p{1cm}<{\centering}p{2.5cm}<{\centering}}
		\toprule
		Event logs & BINet & AE & LSTMAE & BERT & TransformerAE \\
		\midrule
		BPIC2012&	0.5725&	0.5874&	0.9215&	0.9958&	0.9958 \\
		BPIC2012W&	0.5898&	0.7633&	0.9661&	0.9484&	0.9479 \\
		BPIC2013in&	0.5767&	0.6287&	0.8533&	0.9375&	0.9375 \\
		BPIC2017&	0.5636&	0.6778&	0.9842&	0.9997&	0.9997 \\
		\bottomrule
	\end{tabular}
\end{table*}

The comparison with the baseline methods in the anomaly correction task is shown in Fig. \ref{FIG:11}, where the dashed line is the correction result of the trace originally labeled normal. From the figure, it can be seen that the model based on the self-attention mechanism is better than LSTM and fully-connected neural networks in learning sequence context features, and Transformer-AE has the best correction results for anomalous cases when the events anomaly proportion is more than 10\%, and the results of BERT are slightly lower than those of TransformerAE, whereas BERT's results are better than TransformerAE's results when the events anomaly proportion is extremely low, e.g., only have 1 or 2 anomalous events, and the results of both TransformerAE and BERT are better than AE and LSTMAE. For all event logs, AE changes the normal trace greatly, and the result is much lower than the other three models, which also affects the anomaly detection accuracy of the threshold setting method based on average reconstruction loss. That is, AE cannot learn the difference between normal and anomalous trace features well. Because the detection accuracy cannot reach 100\%, the traces that are classified as normal cannot simply be recorded directly into the information system without any changes, but the output of normal traces should be as close as possible to the original traces. In contrast, the other three models all make very small changes to the normal traces, and therefore have low reconstruction errors, meaning that they all learn the differences between normal and anomalous trace features better.

The time spent by each method to process all the test cases in the anomaly detection task is as follows Table \ref{tab 5} shows. From the results, it can be seen that the autoencoder-like methods is much more efficient than the detection method based on time step prediction (BINet), and the models based on the self-attention is also more efficient than the LSTM. In addition TransformerAE has half more Transformer Block than BERT for improving the effect of anomaly correction, so the running time in the anomaly detection task, the two is same.

\begin{table*}[h]
	\caption{Comparison of time spent on each event log in the anomaly detection task (seconds).}
	\label{tab 5}
	\centering
	\begin{tabular}{p{2cm}<{\centering}p{1.2cm}<{\centering}p{1cm}<{\centering}p{1.5cm}<{\centering}p{1cm}<{\centering}p{2.5cm}<{\centering}}
		\toprule
		Event logs & BINet & AE & LSTMAE & BERT & TransformerAE \\
		\midrule
		BPIC2012&	8.6565&	0.0026&	0.1174&	0.0123&	0.0225 \\
		BPIC2012W&	4.8477&	0.0013&	0.0266&	0.0057&	0.0104 \\
		BPIC2013in&	3.1322&	0.0013&	0.0437&	0.0059&	0.0108 \\
		BPIC2017&	19.3421& 0.0055& 0.2633& 0.0260& 0.0456 \\
		\bottomrule
	\end{tabular}
\end{table*}

\section{Conclusion}
\label{sec 6}
In order to achieve efficient detection of anomalies and at the same time correct the anomalies in them. This paper proposes a method for business process anomaly detection and anomaly correction based on Transformer autoencoder by self-supervised learning. Firstly, we create anomalous trace samples through original traces, transform the anomaly detection problem into a classification problem so that we do not need to set the threshold for anomaly detection, utilize the self-attention mechanism to process the complete trace at one time, and can better learn the features of the event sequence context and combine the anomaly detection and anomaly correction tasks based on the structure of the Transformer, and achieve the end-to-end modeling in the anomaly detection and its correction at the same time. Experimental results on four event logs of different sizes show that the proposed method outperforms achieving state-of-the-art in both anomaly detection and anomaly correction tasks. In the future, we will consider event attributes of business process for multi-view business process anomaly detection.

\begin{credits}
\subsubsection{\ackname} This study was funded by the National Natural Science Foundation, China (No. 61572035, 61402011), Anhui Provincial Natural Science Foundation (Water Science Joint Fund, 2308085US11), Key Research and Development Program of Anhui Province (2022a05020005), the Leading Backbone Talent Project in Anhui Province, China (2020-1-12), and Anhui Province Academic and Technical Leader Foundation (No. 2022D327).
\end{credits}

\bibliographystyle{splncs04}
\bibliography{mybibliography}

\end{document}